\documentclass[twocolumn]{galbot}

\usepackage{wrapfig}
\usepackage{tabularx}
\usepackage{textcomp}
\usepackage{stfloats}
\usepackage{url}
\usepackage{verbatim}
\usepackage{graphicx}
\usepackage{titlesec}
\usepackage{tocloft}
\usepackage{adjustbox}
\usepackage{multirow}
\usepackage{tikz}
\usepackage{comment}
\usepackage{amsmath,amssymb} 
\usepackage{colortbl}  
\usepackage{color}
\usepackage{booktabs} 
\usepackage{hyperref}
\usepackage{graphicx}    
\usepackage{subcaption} 
\usepackage{booktabs}
\usepackage{amsmath} 
\usepackage{multirow} 
\usepackage{booktabs} 
\usepackage{arydshln} 
\usepackage{subcaption} 
\RequirePackage{xspace}
\makeatletter
\DeclareRobustCommand\onedot{\futurelet\@let@token\@onedot}
\def\@onedot{\ifx\@let@token.\else.\null\fi\xspace}

\makeatother

\usepackage{makecell}

\definecolor{adptorange}{RGB}{248, 205, 172}
\definecolor{cmpblue}{RGB}{189, 215, 238}
\definecolor{cmpblue}{RGB}{189, 215, 238}

\definecolor{our_red}{RGB}{232,157,160}
\definecolor{our_blue}{RGB}{136,206,230}
\definecolor{our_orange}{RGB}{246,200,168}
\definecolor{our_green}{RGB}{178,211,164}

\definecolor{attn_code0}{RGB}{247,215,200}
\definecolor{attn_code1}{RGB}{238,169,139}
\definecolor{mlp_code0}{RGB}{204,201,221}
\definecolor{mlp_code1}{RGB}{102,95,153}

\definecolor{linecolor1}{RGB}{246, 248, 239}
\definecolor{linecolor2}{RGB}{230, 234, 217}
\definecolor{linecolor3}{RGB}{211, 222, 190}

\definecolor{token_blue}{RGB}{84, 120, 140}

\definecolor{myMagenta}{rgb}{0.9,0,0.4}

\usepackage{pifont}       
\usepackage{bbding}       
\usepackage{fontawesome}
\usepackage{xspace}

\usepackage{float}

\newlength\savewidth

\newcolumntype{x}[1]{>{\centering\arraybackslash}p{#1pt}}
\newcolumntype{y}[1]{>{\raggedright\arraybackslash}p{#1pt}}
\newcolumntype{z}[1]{>{\raggedleft\arraybackslash}p{#1pt}}

\renewcommand{\paragraph}[1]{\vspace{1mm}\noindent\textbf{#1}}
\usepackage{colortbl}
\usepackage{xcolor}
\usepackage{wrapfig}
\usepackage{graphicx}
\usepackage{amssymb}
\usepackage{adjustbox}
\usepackage{colortbl}
\usepackage{xcolor}
\usepackage{multirow}
\usepackage{array}
\usepackage{makecell}
\usepackage{bbm}
\usepackage{collcell,xfp}
\usepackage{pgf}
\usepackage{tikz}
\usepackage[most]{tcolorbox}
\usepackage{csquotes}
\usepackage[noorphans,vskip=1em,leftmargin=1em]{quoting}
\usepackage{enumitem} 
\usepackage{tcolorbox} 
\usepackage{forest}
\usepackage{wrapfig}
\usepackage{caption}
\usepackage{subcaption}
\usepackage{longtable}
\usepackage[T1]{fontenc}

\renewcommand{\paragraph}[1]{\vspace{1.25mm}\noindent\textbf{#1}}

\usepackage{algorithm}
\usepackage{listings}

\definecolor{codeblue}{rgb}{0.25, 0.5, 0.5}
\definecolor{codekw}{rgb}{0.35, 0.35, 0.75}
\lstdefinestyle{Pytorch}{
    language = Python,
    backgroundcolor = \color{white},
    basicstyle = \fontsize{9pt}{8pt}\selectfont\ttfamily\bfseries,
    columns = fullflexible,
    aboveskip=1pt,
    belowskip=1pt,
    breaklines = true,
    captionpos = b,
    commentstyle = \color{codeblue},
    keywordstyle = \color{codekw},
}

\definecolor{green}{HTML}{009000}
\definecolor{red}{HTML}{ea4335}

\newcommand{\infobox}[1]{
    \vspace{-0.18cm}
    \begin{tcolorbox}[
        colback=white!90!gray,     
        colframe=teal!60!black,   
        arc=5pt,                   
        boxsep=5pt,                 
        left=5pt,                  
        right=10pt,                 
        top=2pt,                   
        bottom=3pt,                
        boxrule=0.8pt,              
        drop shadow=gray!50!white, 
        enhanced jigsaw             
    ]
    \vspace{-0.1cm}
         \textit{#1}
    \vspace{-0.2cm}
    \end{tcolorbox}
    \vspace{-0.15cm}
}

\newcommand{\methodname}{Humanoid-GPT\xspace}

\title{Humanoid-GPT: Scaling Data and Structure for Zero-Shot Motion Tracking}

\author[1, 2, *]{Zekun Qi}
\author[1, 2, *]{Xuchuan Chen}
\author[2, *]{Dairu Liu}
\author[2, *]{Chenghuai Lin}
\author[1, 2]{Yunrui Lian}
\author[2, 3]{Sikai Liang}
\author[1, 2]{Zhikai Zhang}
\author[1, 2]{Yu Guan}
\author[2]{Jilong Wang}
\author[2, 3]{Wenyao Zhang}
\author[2]{Xinqiang Yu}
\author[2, 4, \dagger]{He Wang}
\author[1, 5, \dagger]{Li Yi}

\affiliation[1]{Tsinghua University}
\affiliation[2]{Galbot Inc.}
\affiliation[3]{Shanghai Jiao Tong University}
\affiliation[4]{Peking University}
\affiliation[5]{Shanghai Qi Zhi Institute}

\contribution[*]{Equal Contribution}
\contribution[\dagger]{Corresponding author}

\date{\today} 
\code{\url{https://github.com/GalaxyGeneralRobotics/Humanoid-GPT/}}
\page{\url{https://qizekun.github.io/Humanoid-GPT/}}

\abstract{
We introduce \textbf{Humanoid-GPT}, a GPT-style \textbf{Transformer} with \textbf{causal attention} trained
on a billion-scale motion corpus for whole-body control. Unlike prior shallow MLP trackers constrained by scarce data and an agility–generalization trade-off, Humanoid-GPT is pre-trained on a \textbf{2B-frame} retargeted corpus that unifies all major mocap datasets with large-scale in-house recordings. Scaling both data and model capacity yields a single generative Transformer that tracks highly dynamic behaviors while achieving unprecedented \textbf{zero-shot} generalization to unseen motions and control tasks. Extensive experiments and scaling analyses show that our model establishes a new performance frontier, demonstrating robust zero-shot generalization to unseen tasks while simultaneously tracking highly dynamic and complex motions.
}

\makeatletter
\apptocmd{\mymaketitle}{%
  \vspace{15pt}
  \begin{center}
      \captionsetup{type=figure}
      \includegraphics[width=\textwidth]{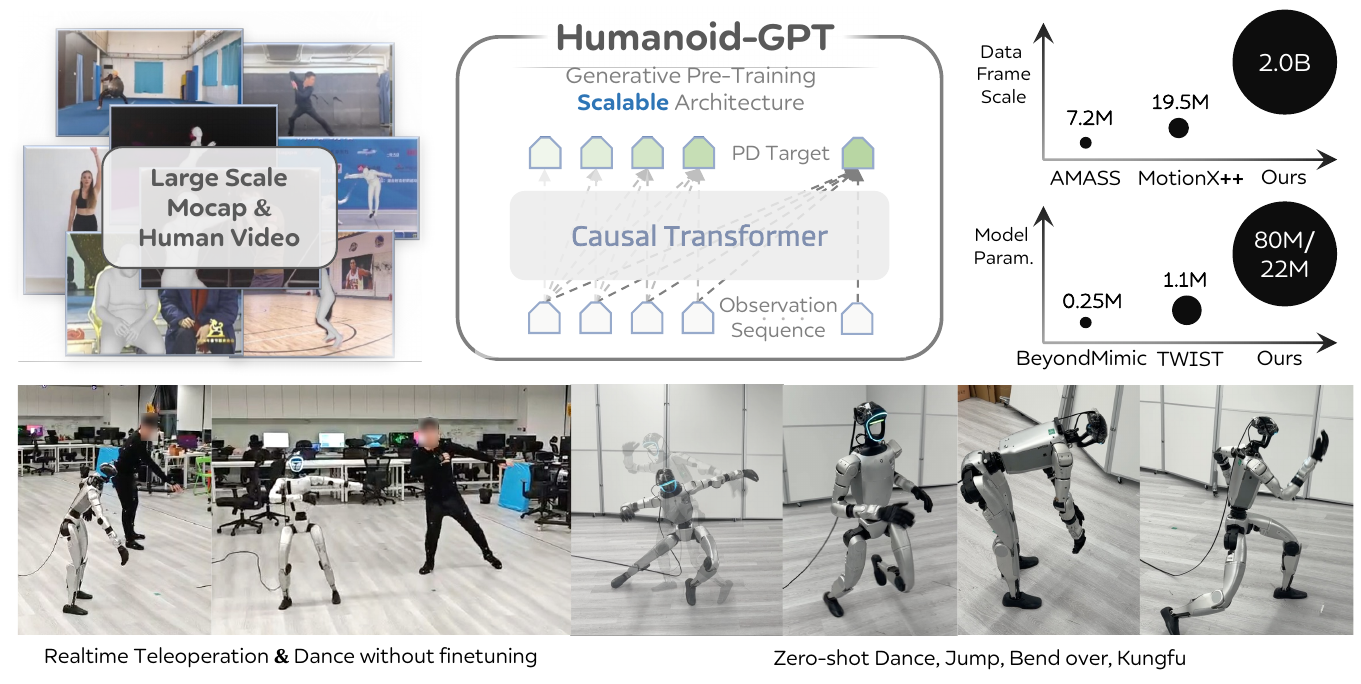}
      \vspace{1pt}
      \captionof{figure}{\textbf{Overview of Humanoid Generative Pre-Training for Zero-Shot Motion Tracking.}
      We introduce Humanoid-GPT, a Transformer-based humanoid tracker trained on an unprecedented 2B-frame retargeted corpus, whose scaling in data and model capacity yield a universal controller that zero-shot tracks highly dynamic and unseen human motions beyond the agility and generalization limits of prior MLP trackers. The figure illustrates the real-time whole-body control and zero-shot dance generation capabilities of Humanoid-GPT.}
      \label{fig:teaser}
      \vspace{5pt}
  \end{center}
}{}{}
\makeatother

\begin{document}
\maketitle
\pagestyle{empty}

\section{Introduction}
\label{sec:intro}

Artificial General Intelligence (AGI) for embodied agents is ultimately a \emph{generalization} problem: a humanoid should execute robust whole-body behaviors under unseen tasks, styles, and environments~\cite{ChatGPT22,GPT3_20,GPT4o24,gpt_o1}. In language and vision, the most reliable path to generalization has been \emph{scale}---larger data, larger models, and carefully designed training objectives~\cite{ChatGPT22,InstructGPT22,gemini23,gpt_o1,LLaMA23,CLIP21,SAM23}. Scaling is not only a recipe for better average performance; it often unlocks new capabilities and predictable trends~\cite{emergent22}.

Humanoid motion tracking has not followed this trajectory. Current trackers are typically shallow MLPs trained on small motion corpora. Even widely used datasets~\cite{amass19,lafan20,omomo23} contain only on the order of $10^4$ trajectories (about $7.2$M frames). This mismatch in scale creates a persistent failure mode: \emph{agility and generalization trade off}. Trackers that excel on in-domain agile motions often break on unseen styles, while trackers that generalize modestly tend to underfit complex dynamics and lose sharpness in tracking. Recent results make this tension clear: BeyondMimic~\cite{beyondmimic25} and ASAP~\cite{asap25} track agile motions well but do not generalize zero-shot to unseen movements; TWIST~\cite{twist25} and UniTracker~\cite{unitracker25} generalize better but struggle on highly dynamic actions.

We argue that this trade-off is not fundamental. It is a symptom of \emph{insufficient scale} and \emph{mismatched training design}. Simply adding more motion clips to the same pipeline is not enough. When the scale increases by orders of magnitude, three questions become decisive:
\ding{182} \textbf{What data} should we train on, and how do we process the large, noisy data?
\ding{183} \textbf{What model structure} matches the online tracking constraint and continues to improve with scale?
\ding{184} \textbf{What training recipe} remains stable when the dataset grows from millions to billions of frames?

This paper answers these questions and presents \textbf{Humanoid-GPT}, a universal, online humanoid motion tracker built around the science of scaling.

\paragraph{Science of Scale.}
We construct a motion corpus at a new regime for tracking. We aggregate all widely available mocap sources, including Lafan1~\cite{lafan20}, AMASS~\cite{amass19}, Motion-X++~\cite{motionx++25}, PHUMA~\cite{phuma25}, and MotionMillion~\cite{motionmillion25}, and we add a large internally captured dataset for real-world coverage. After strict filtering, segmentation, and augmentation, we obtain \textbf{2B G1-retargeted motion frames/tokens}, over \textbf{$200\times$} larger than prior tracker training sets. This scale forces changes that smaller systems can ignore: we redesign key reward components and re-tune sensitive hyperparameters to keep training stable. Crucially, we provide the first systematic evidence that \emph{video-estimated motion} can materially improve tracking when the model and the training set are scaled appropriately.

\paragraph{Modern Structure for Online Tracking.}
Motion tracking for control is inherently \emph{causal}: at test time the policy cannot access future observations. Many existing trackers still rely on non-causal modeling choices or capacity-limited MLPs. We instead adopt a \textbf{scalable Transformer} with GPT-style causal attention. The model predicts per-joint PD targets with causal temporal attention, which aligns with the deployment constraint by design. This structure also scales cleanly with data and model size, unlike shallow MLPs and non-causal variants that saturate early.

\begin{table}[t]
  \centering
  \caption{\textbf{Comparison of \methodname with related works.}}
  \label{tab:comp}
  \resizebox{1.0\linewidth}{!}{
  \setlength{\tabcolsep}{1.5pt}
  \scriptsize
  \begin{tabular}{lcccc}
    \toprule
    \textbf{Method} & \textbf{Low-level Tracker} & \textbf{Agile} & \textbf{Zero-shot} & \textbf{\#Frames} \\
    \midrule
    HumanPlus~\cite{humanplus24}    & Transformer & $\times$ & $\times$ & 7.2M \\
    OmniH2O~\cite{omnih2o24}        & MLP & $\times$ & $\times$ & 7.2M \\
    ASAP~\cite{asap25}              & MLP & \checkmark & $\times$ & - \\
    GMT~\cite{gmt25}                & MoE-MLP & \checkmark & $\times$ & 6.0M \\
    UniTracker~\cite{unitracker25}  & MLP & \checkmark & $\times$ & 7.2M \\
    BumbleBee~\cite{BumbleBee25}  & Transformer & \checkmark & $\times$ & 7.2M \\
    TWIST~\cite{twist25}            & MLP & $\times$ & $\sim$ & 9.2M \\
    Any2Track~\cite{any2track25}    & MLP & \checkmark & $\times$ & 9.1M \\
    SONIC~\cite{sonic25}    & MLP & \checkmark & \checkmark & 100M \\
    \textbf{\methodname (ours)}      & Transformer & \checkmark & \checkmark & \textbf{2.0B} \\
    \bottomrule
  \end{tabular}
  }
  \vspace{-10pt}
\end{table}

\paragraph{Balanced Diversity Matters.}
More data does not automatically mean better generalization. In large motion corpora, common styles dominate and rare but important behaviors vanish in the long tail. We introduce \textbf{Harmonic Motion Embedding (HME)} as a representation learning tool that measures and organizes motion diversity directly from raw motion. HME enables \textbf{diversity-aware, distribution-balanced sampling} during training. Our analysis shows a simple but powerful insight: \textbf{diversity and balance are both necessary}. Diversity without balance still overfits frequent modes; balance without diversity caps capability.

\paragraph{Results and scaling laws.}
With these ingredients, Humanoid-GPT substantially improves \emph{both} agility and zero-shot generalization. We further derive a \textbf{scaling law} for humanoid motion tracking that relates performance to data scale and model capacity, offering a concrete roadmap for future general-purpose whole-body control.

Compared with prior work that either applies Transformer controllers on limited motion hours~\cite{humanplus24} or scales MLP-based policies on hundreds of millions of frames~\cite{sonic25}, Humanoid-GPT is, to our knowledge, the first system that (i) distills a large library of RL motion experts into a single GPT-style tracker, (ii) trains on a curated \textbf{2B-frame} corpus, and (iii) systematically characterizes how \textbf{data scale, model scale, and diversity balance} jointly govern zero-shot agile motion tracking on real humanoid hardware.

\section{Related work}
\label{sec:related_work}

\subsection{Large-scale Motion Data}

Large-scale motion datasets have become essential for learning generalizable human motion tracking. Early datasets \citep{lafan20,amass19,omomo23} offered high-quality but studio-constrained motions, limiting diversity. With video-based reconstruction and large-scale synthetic generation, recent datasets greatly expand motion coverage, incorporating diverse activities, styles, and subjects with multimodal supervision \citep{motionx23, motionx++25,motionmillion25,humanoidx24}.
More recently, \citep{phuma25} provides physically consistent motions with contact modeling, joint constraints, and reduced foot-sliding, offering stability benefits over purely kinematic sources.
Together, these increasingly diverse and physically grounded datasets supply richer motion variety and stronger physical priors, forming key foundations for unified and robust human motion tracking systems.


\begin{figure*}[t!]
  \centering
  \includegraphics[width=1.0\linewidth]{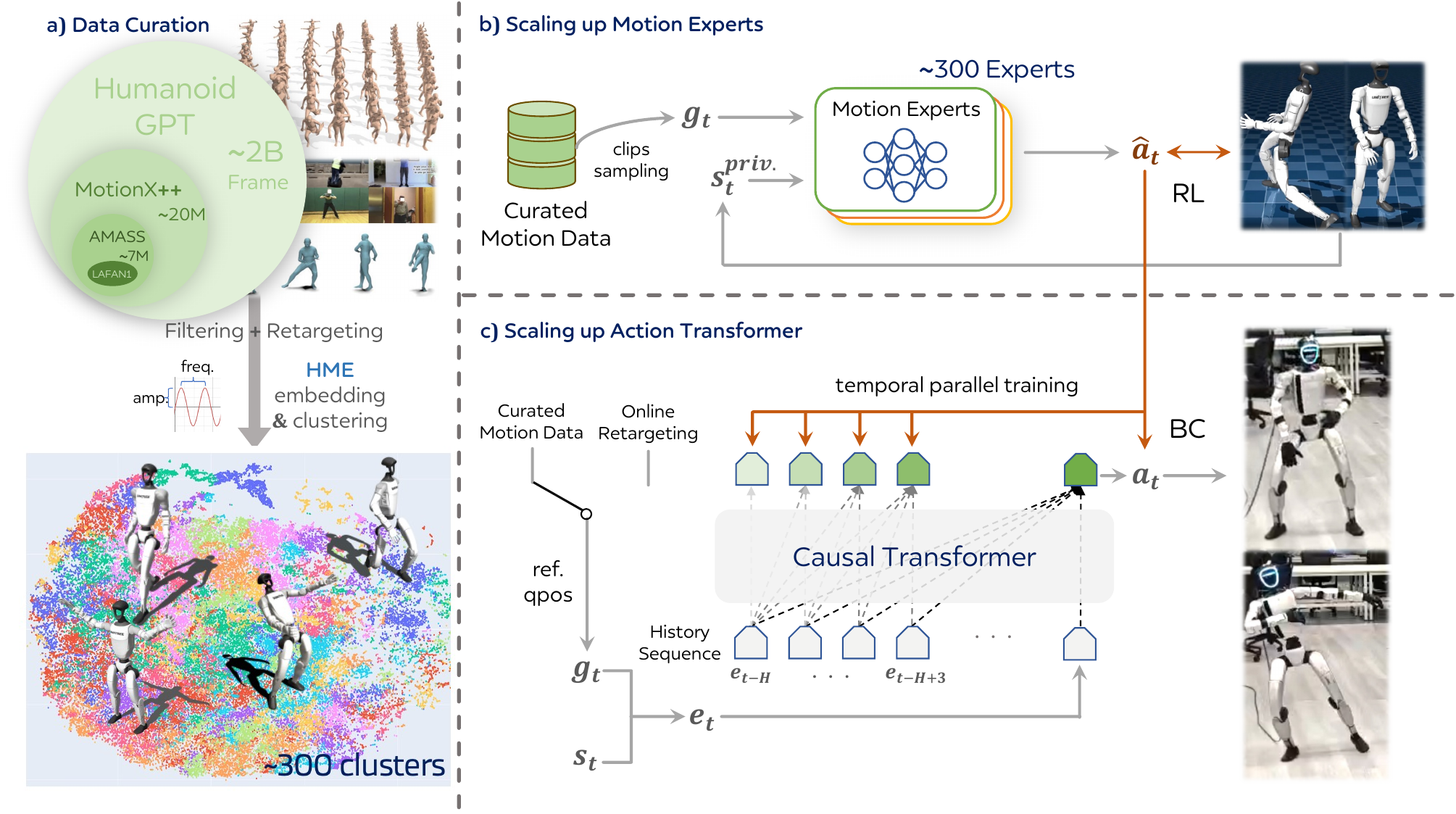}
  \vspace{2pt}
  \caption{
   \textbf{Overview of \methodname}. 
  The system consists of three stages: (\textbf{a}) data curation and processing, 
  (\textbf{b}) training PPO-based motion experts on clusters with keypoint-level rewards, 
  and (\textbf{c}) distilling all experts into a single Transformer-based generalist policy via parallel DAgger supervision. The resulting \methodname can take unseen or online retargeted motions as reference inputs and track them in a fully zero-shot manner.
  }
  \label{fig:pipeline}
\end{figure*}

\subsection{Learning Human Motion Tracking}
Physics-based tracking aims to produce temporally coherent, dynamically feasible whole-body control from reference motions. Early works~\cite{phc23,uhm23,synchronized23,freemotion24} establish the paradigm of coupling imitation with contact-aware stability in simulation, while subsequent pipelines~\cite{humanplus24,omnih2o24,exbody24,exbody2_24,omniretarget25,kungfubot25,clone25,anyteleop23,twist25,R2S2_25} extend to real-world deployment on specific platforms.

Recent efforts shift toward generalization. GMT~\cite{gmt25} employs Mixture-of-Experts with adaptive sampling; UniTracker~\cite{unitracker25} adopts a CVAE-based teacher-student framework---both improving coverage but remaining constrained by limited motion scale. SONIC~\cite{sonic25} scales to 100M frames with an MLP controller, yet MLP capacity saturates as data grows. HumanPlus~\cite{humanplus24} introduces a Transformer controller but trains it with standard PPO, missing the parallelism advantage inherent to Transformers.

As summarized in Table~\ref{tab:comp}, existing methods either rely on curated motion sets or architectures that do not scale gracefully. \methodname~reframes tracking as GPT-style sequence modeling: we distill hundreds of RL experts into a causal Transformer trained on 2B frames, achieving strong zero-shot generalization where similarly-sized MLPs plateau.

\section{Scaling up Humanoid Motion Data}
\label{sec:methods}

We first collect and curate large-scale human motion data to ensure the fidelity of motion dynamics, followed by retargeting these motions to the humanoid joint space.

\subsection{Data Curation}
\label{subsec:method_data}
Constructing a high-quality motion dataset is essential for ensuring the fidelity and diversity of motion dynamics in zero-shot humanoid motion tracking. Existing datasets~\cite{amass19, lafan20, motionx23, humanoidx24} often contain limited categories of motion capture sequences or exhibit inconsistencies in physical plausibility and spatial alignment, which constrain their generalization to complex whole-body tracking scenarios. Recent advances in large-scale motion generation ~\cite{motionmillion25} and physically grounded motion filtering ~\cite{phuma25} have introduced abundant and high-quality motion priors, significantly extending the coverage of motion distributions. To fully exploit these available resources, we curate a large-scale motion corpus by aggregating AMASS~\cite{amass19}, LAFAN1~\cite{lafan20}, MotionMillion~\cite{motionmillion25}, and PHUMA~\cite{phuma25}, shown in Fig.~\ref{fig:pipeline}\textbf{(a)}, encompassing a wide spectrum of human activities that serve as the foundation for \methodname.

After assembling the datasets into a unified corpus, we employ an off-the-shelf motion retargeting framework~\cite{gmr25} to map each human motion sequence into the 29-DoFs joint space of the Unitree-G1 humanoid. During this process, we further filter out sequences involving explicit object interactions—such as sitting on chairs, swimming, or stair climbing—to ensure the resulting motions are compatible with the humanoid’s actuation in a plain scene. To further enrich temporal variability and improve robustness to motion speed, we apply motion time-warping augmentation by uniformly accelerating and decelerating every sequence, expanding the dataset to approximately five times its original size. This yields a clean, physically consistent, and diverse dataset suitable for downstream reinforcement-learning based expert training.

\subsection{Harmonic Motion Embedding}

To balance the trade-off between motion coverage and training efficiency, we partition the full motion corpus into multiple clusters and train each expert on a specific motion subset. To cluster motions directly in the latent space, we propose a novel embedding representation called \textbf{Harmonic Motion Embedding (HME)}. Concretely, we first train several Periodic Autoencoders~\cite{pae22} on different data partitions to extract per-joint periodic amplitudes and frequencies from each motion sequence. For each sequence, we then aggregate the mean and standard deviation of these joint-level harmonic features to obtain its HME vector, yielding a compact and descriptive embedding for the entire corpus. Finally, we apply K-Means clustering over all HME embeddings using pairwise distances as the similarity metric, producing roughly 300 motion clusters. Each cluster contains about 1k–2k sequences, offering strong intra-cluster consistency while preserving broad coverage of the global motion distribution.

\section{Scalable Generative Tracker}

We present the \textbf{\methodname} framework. Built via a two-stage pipeline: RL-trained motion experts followed by transformer distillation, our model enables humanoids to track arbitrary human motions without any finetuning.

\subsection{Training Motion Experts}
To enable diverse motion priors for the \methodname, we train multiple motion experts to collectively cover the dynamic distribution present in the dataset. 

On each cluster, we train a PPO-based policy to track all the sequences inside the cluster, presenting in Fig \ref{fig:pipeline}\textbf{(b)}. The policy is formulated as $\pi: \mathcal{G} \times \mathcal{S} \mapsto \mathcal{A}$, which maps the input reference joints and humanoid proprioceptive observations to low-level motor actions. At each time step $t$, the policy receives the current privileged robot state $s_t^{priv.}$ along with the target reference pose $q_t^{\text{ref}}$ extracted from the motion clip. The state $s_t^{priv.}$ encodes per-joint positions and velocities, the root’s angular velocity, projected gravity, and the previous control action. The policy outputs per-joint actions $a_t$, which are converted into actuator torques through a PD controller. 
The motion tracking objective is to drive the robot's state to match the 
target pose $g_t = q_t^{\text{ref}}$ while maintaining balance and dynamic stability. 
To enforce physically grounded tracking, the reward is computed at the body keypoint level, including position and velocity consistency terms for critical parts of the body (e.g., arms, hips, feet, pelvis). Let $\mathcal{K}$ denote the set of tracked body keypoints. For each keypoint $k\!\in\!\mathcal{K}$ at time $t$, let $e^{\text{pos}}_{k,t}\!\in\!\mathbb{R}^3$ and $e^{\text{vel}}_{k,t}\!\in\!\mathbb{R}^3$ denote the position and velocity residuals between the humanoid and the reference motion, and let $\theta_{k,t}$ be the rotation error induced by the $\mathrm{SO}(3)$ log map.  With positive keypoint weights $w_k$ and scaling factors 
$\alpha_{\text{pos}}, \alpha_{\text{rot}}, \alpha_{\text{vel}}$,  
the abstract keypoint reward is formulated as
\begin{equation}
\begin{aligned}
R_{\text{kpt}}(t) &= R_{\text{pos}}(t) + R_{\text{rot}}(t) + R_{\text{vel}}(t) + R_{\text{penal}}(t),\\
R_{\text{pos}}(t) &= \sum_{k\in\mathcal{K}} w_k \exp\!\left(-\alpha_{\text{pos}}\|e^{\text{pos}}_{k,t}\|_{1}\right),\\
R_{\text{rot}}(t) &= \sum_{k\in\mathcal{K}} w_k \exp\!\left(-\alpha_{\text{rot}}\theta_{k,t}\right),\\
R_{\text{vel}}(t) &= \sum_{k\in\mathcal{K}} w_k \exp\!\left(-\alpha_{\text{vel}}\|e^{\text{vel}}_{k,t}\|_{1}\right).
\end{aligned}
\end{equation}
The exponential form softly penalizes deviations in position, orientation, 
and velocity across all body keypoints, and $R_{\text{panel}}(t)$ consists of several penalties like self-contacts and smoothness, promoting a globally accurate yet locally stable motion tracking.

During training, we randomly sample short motion clips as tracking targets and evaluate each expert using root pose error, velocity error, and stable tracking duration. These metrics ensure convergence toward physically consistent motion reproduction within each cluster. After training, only experts that achieve high-fidelity and long-horizon stability are retained, forming a diverse library of motion priors that provides \methodname\ with a physically grounded initialization across heterogeneous motion regimes.

\begin{figure}[t]
  \centering
  \includegraphics[width=1.0\linewidth]{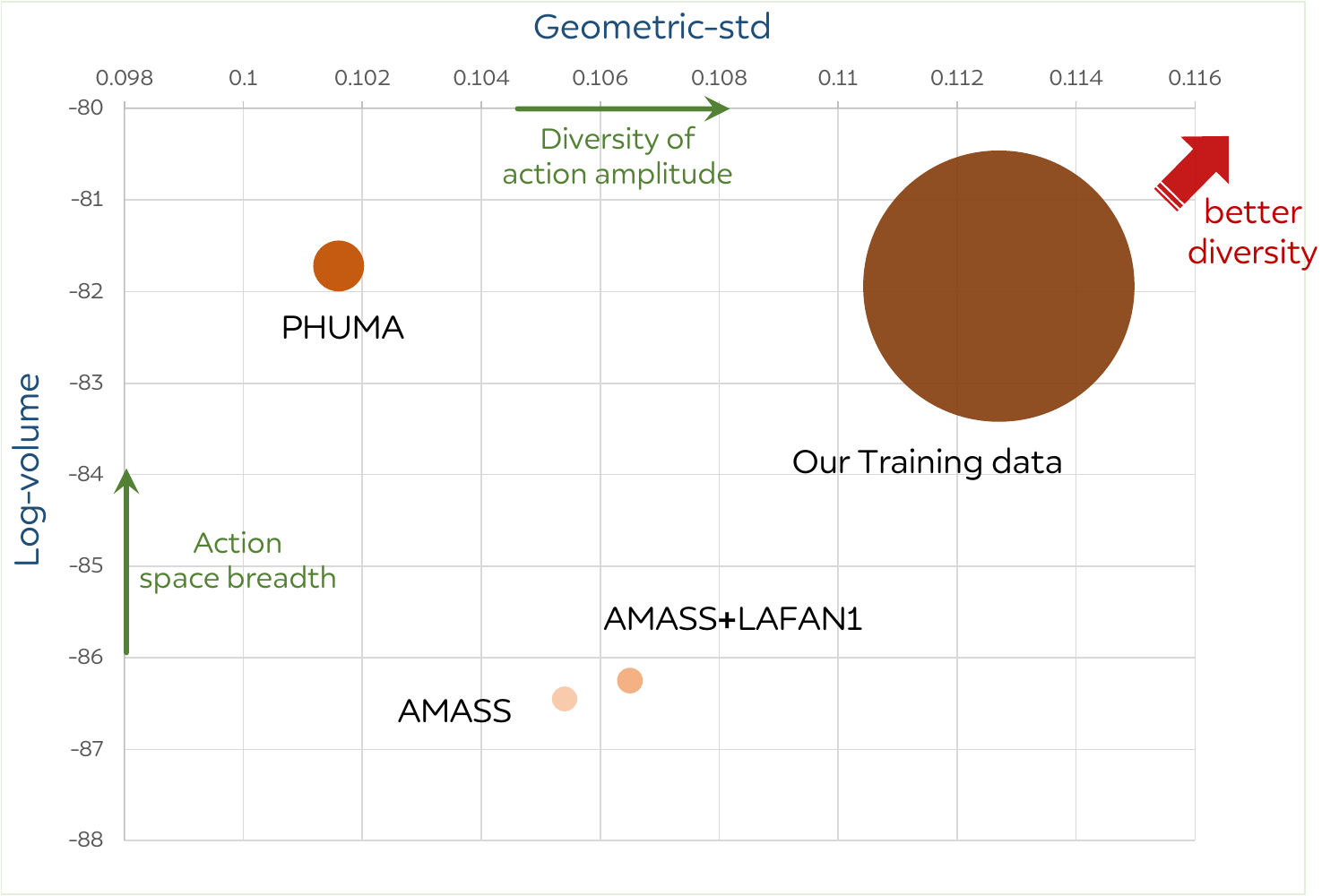}
  \caption{
  \textbf{Comparison of dataset diversity in the HME embedding space.}
  Each bubble represents a dataset, where the horizontal and vertical axes denote \textit{gstd} and \textit{log-volume} respectively, and the bubble size reflects the relative amount of motion clips. Upper-right bubbles indicate broader coverage and higher diversity.
  }
  \label{fig:diversity_bubble}
\end{figure}

\subsection{Building Zero-shot Foundational Tracker}
The motion experts trained above can accurately reproduce physically-grounded motions within their own clusters, but tend to degrade sharply when encountering out-of-distribution motion targets. To bridge the gaps across motion domains and consolidate their specialized knowledge, we introduce a distillation stage illustrated in Fig.\ref{fig:pipeline}\textbf{(c)} that transfers the behaviors of all experts $\mathcal{T}$ into a single unified policy, adopting the DAgger~\cite{dagger11} framework to distill the knowledge of all motion experts into a single generalist policy. 

To distill expert behaviors efficiently, we reformulate the distillation process 
as a sequence modeling problem and employ a Transformer~\cite{transformer17}-based generalist tracker $G_\theta$. At each timestep $t$, the input token embedding $e_t$ is constructed by concatenating the current proprioceptive state $s_t$ and the target reference pose $q_{t}^{\text{ref}}$ from the motion clip. A sequence of length $H$ containing such tokens $\{e_{t-H+1}, e_{t-H+2}, \dots, e_t\}$ is fed into the Transformer with a temporal causal mask, allowing the model to capture long-horizon dependencies and temporal consistency across the trajectory. After a forward pass, actions at all output positions will be supervised by the corresponding history of teacher $t_i$'s output, empowering the model to be trained on DAgger feedback efficiently over multiple timesteps in a single pass, shown in eq.\eqref{eq:loss}. We use SmoothL1Loss as our loss function $\mathcal{L}$.

\begin{equation}
\label{eq:loss}
\begin{aligned}
    \hat{a}_{t-H+1:t} = \bigcup_{t_i\in\mathcal{T}} \operatorname*{concat}_{k\in[-H+1, 0]} t_i(s_{t-k}^{priv.}, g_{t-k})\\
    l = \mathcal{L}(G_\theta(e_{t-H+1:t}), \hat{a}_{t-H+1:t})
\end{aligned}
\end{equation}

During inference, we maintain a queue of maximal $H$ history tokens as the input of transformer and use the output located in the last position as the current control target. 

This design of our \methodname model naturally exploits the Transformer’s inherent strengths of parallel sequence supervision and autoregressive temporal predicting,  Moreover, because tokens at different positions attend to varying amounts of historical context, the trained model implicitly learns position-invariant temporal prediction, enabling it to output stable and physically consistent control targets even at the beginning of an episode, where historical information is scarce.

\begin{figure*}[t!]
  \centering
  \vspace{-10pt}
  \includegraphics[width=1.0\linewidth]{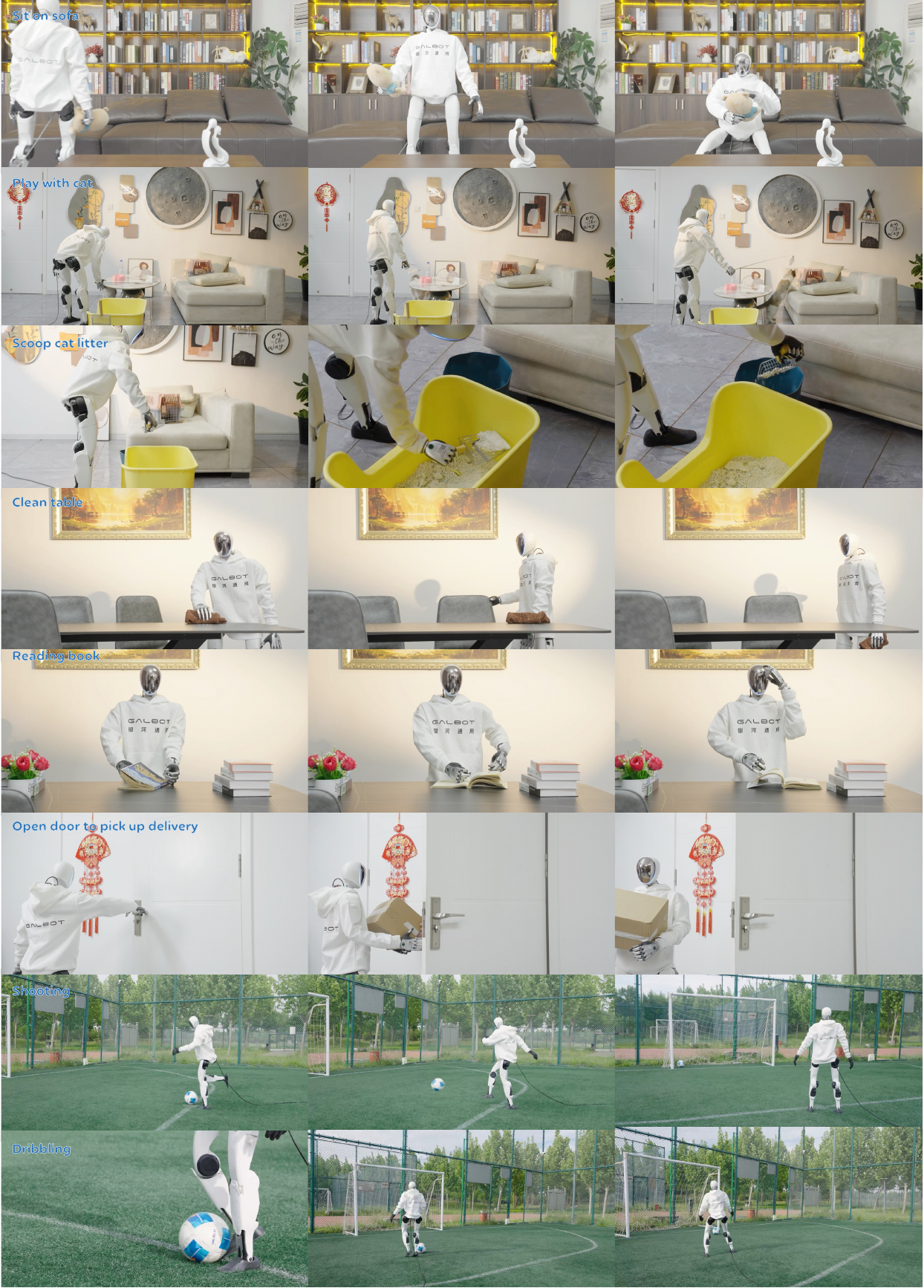}
  \caption{\textbf{Real-world experiments for our \methodname.} All motions illustrated are excluded from training to verify generalization capability. Our method can track diverse, complex and high-dynamic motion in a zero-shot manner.}
  \vspace{-15pt}
  \label{fig:real_exp}
\end{figure*}

\section{Experiments}
\label{sec:experiments}

In this section, we aim to answer the following core questions that arise from our exploration of the scaled-up humanoid motion tracker and the emergence of zero-shot generalization in \methodname:

\begin{itemize}[leftmargin=0em]

\vspace{5pt}
\infobox{Scaling Law in Humanoid Tracking.} 
\vspace{5pt}
How does the zero-shot tracking ability of \methodname\ scale with the amount of training data and model capacity? 
Does enlarging the motion corpus or Transformer parameters yield predictable improvements in stability and generalization?
  
\vspace{5pt}
\infobox{Diversity and Balance Drives Generalization.}
\vspace{5pt}
How can we quantify the diversity of motion data, and how does such diversity contribute to zero-shot tracking performance? 

\vspace{5pt}
\infobox{Powerful Structure Matters.}
\vspace{5pt}
How does the architectural choice, such as the Transformer versus other alternatives, affect the model's ability to capture long-horizon dynamics and generalize across unseen motions?
\end{itemize}

We systematically design experiments to address these questions, analyzing scaling trends, diversity–generalization relationships, and architectural contributions to robust zero-shot humanoid tracking.

\begin{table*}[t]
  \centering
  \caption{
  Comparison of backbone architectures and scaling effects. 
  Larger datasets and higher-capacity Transformers consistently improve stability 
  and zero-shot tracking accuracy across all metrics.
  }
  \label{tab:sim_scaling_backbone}
  \resizebox{1.0\linewidth}{!}{
  \setlength{\tabcolsep}{5.5pt}
  \footnotesize
  \begin{tabular}{cccccccc}
    \toprule
    Backbone &
    \#Train Tokens &
    \#Model Params.(M) &
    SR $\uparrow$ &
    MPJPE $\downarrow$ &
    MPJVE $\downarrow$ &
    RootVelErr $\downarrow$ &
    MPKPE $\downarrow$ \\
    \midrule
    MLP (3-layer)   & 2M  & 0.25M & 76.89 & 0.1191 & 0.6081 & 0.2304 & 100.49 \\
    TCN (8-layer)   & 2M  & 0.65M & 81.48 & 0.0885 & 0.5716 & 0.2266 & 79.75 \\
    \midrule
    Humanoid-GPT-S   & 2M  & 5.7M & 83.26 & 0.0853 & 0.5492 & 0.2049  & 62.65 \\
    Humanoid-GPT-S  & 20M  & 5.7M & 86.02 & 0.0802 & 0.5210 & 0.1868  & 46.49 \\
    Humanoid-GPT-B  & 200M & 22.1M & 88.27 & 0.0793 & 0.5076 & 0.1820  & 44.78 \\
    \rowcolor{linecolor1}Humanoid-GPT-B & 2B  & 22.1M & 90.43 & 0.0768 & 0.4891 & \textbf{0.1756}  & 41.49 \\
    \rowcolor{linecolor2}Humanoid-GPT-L & 2B  & 80.4M & \textbf{92.58} & \textbf{0.0735} & \textbf{0.4820} & 0.1785  & \textbf{40.99} \\
    \bottomrule
  \end{tabular}
  }
\end{table*}

\subsection{Experiment Settings}

We evaluate our method in both simulation and real-world settings, 
using the 29-DoF Unitree-G1 as the humanoid platform for tracking the target motion in all experiments. In simulation, we employ MuJoCo\citep{mujoco12} as the physics engine to quantitatively assess the performance of different data and model variants under controlled conditions. For real-world evaluation, we adopt an online motion-retargeting pipeline that continuously converts the motion of a MoCap actor into the G1’s joint space, which serves as the online reference trajectories for our \methodname\ to track. 

\subsection{Analysis on Data Diversity}
We first evaluate how the diversity of motion data influences the generalization ability of our tracker. We compare three datasets with an increasing amount of clips: the commonly used \textbf{AMASS}~\citep{amass19}, the extended \textbf{AMASS+LAFAN}~\citep{lafan20}, recent \textbf{PHUMA}\citep{phuma25}, and our curated large-scale dataset described in Sec.~\ref{subsec:method_data}, which aggregates AMASS, LAFAN1, MotionMillion, and PHUMA to cover a substantially broader range of human dynamics. To quantitatively measure dataset diversity in the latent space, we compute the geometric mean standard deviation (\textit{gstd}) and the logarithmic volume of the covariance ellipsoid (\textit{log-volume}) based on the HME embeddings introduced in Sec~\ref{subsec:method_data}. To ensure that the size of each dataset does not bias the diversity estimation, we uniformly sample 10{,}000 embeddings from each dataset for evaluation. Given an embedding matrix $X = [x_1, x_2, \dots, x_N]^\top \in \mathbb{R}^{N \times D}$, we compute the covariance $\Sigma$ and the two diversity indicators are then defined as

\begin{equation}
\begin{aligned}
    \text{gstd} &= \exp\!\left( \frac{1}{D} \sum_{j=1}^{D} \log \sigma_j \right), \\
    \text{log-volume} &= \tfrac{1}{2}\log\det(\Sigma + \epsilon I),
\end{aligned}
\end{equation}
where $\sigma_j$ denotes the standard deviation of the $j$-th embedding dimension and $\epsilon$ is a small regularization term ensuring numerical stability. Higher values indicate that the dataset spans a larger and more uniformly distributed region in the latent manifold. As shown in Fig.~\ref{fig:diversity_bubble}, our curated dataset exhibits both higher embedding scale and broader latent coverage, with an approximately $4\!-\!5$ increase in log-volume compared with AMASS. This result highlights that richer motion diversity substantially expands the latent coverage of the motion manifold, providing stronger priors for robust zero-shot humanoid tracking.

\subsection{Evaluation in Simulation}
We first evaluate \methodname\ in the MuJoCo simulation to systematically analyze the effects of data and model scaling on zero-shot tracking performance. 
This controlled setup allows us to precisely measure stability, fidelity, 
and generalization across diverse motion categories before transferring to the real world.

\paragraph{Setup.}
We construct multiple training configurations by varying both the size of the motion corpus and the capacity of the Transformer backbone. 
Specifically, we sample teachers from clusters of 10k, 50k, 100k, and 300k motion clips for data scaling, and employ Transformer models with different parameters for model scaling, resulting in \methodname-small, \methodname, \methodname-Large. Each configuration is trained with identical DAgger distillation settings to ensure fair comparison. We test all variants above in the \textbf{AMASS-test} split in \citep{amass19} following \citep{phc23}, which is an unseen subset during training.

\paragraph{Compared methods.}
We compare the following baseline policies, which represent the strongest publicly available humanoid trackers at the time of writing and are all based on MLP-style low-level controllers trained on around $6$--$9$M motion frames (see Tab.~\ref{tab:comp}):
\begin{itemize}[leftmargin=1.0em]
    \item \textbf{GMT}~\cite{gmt25}: A Mixture-of-Experts (MoE) tracker trained on a subset of AMASS~\cite{amass19} motions, where each expert specializes in a particular motion pattern and the gating network selects appropriate experts to maintain physically consistent whole-body tracking.
    \item \textbf{TWIST}~\cite{twist25}: A whole-body imitation policy distilled from the TWIST teleoperation system, designed for responsive human-in-the-loop control on Unitree humanoids and trained on a large corpus of teleoperated demonstrations covering everyday and dynamic behaviors.
    \item \textbf{Any2Track}~\cite{any2track25}: A general tracker trained on AMASS~\cite{amass19} and LAFAN1~\cite{lafan20}, which emphasizes robustness to perturbations by incorporating dynamics-adaptive control objectives and strong disturbance randomization during training.
\end{itemize}
For all three methods, we use the authors' released implementations and checkpoints and evaluate them under the same simulation and retargeting protocol as our Humanoid-GPT models to ensure a fair comparison.

\paragraph{Metrics.}
We report three quantitative metrics: \ding{182}: \textbf{Tracking Success Rate (SR)}, which measures the proportion of trajectories 
that can be stably tracked without falling. \ding{183}: \textbf{Mean per-Joint Position Error (MPJPE)} (\textit{rad}) as the average position error of all joints. \ding{184}: \textbf{Mean per-Joint Velocity Error (MPJVE)} (\textit{rad/s}) as the average angular velocity error of all joints. \ding{185}: \textbf{Root Velocity Error (RootVelErr)} (\textit{m/s}) as the average linear velocity error of the humanoid's base, and \ding{186}: \textbf{Mean per-Keypoint Position Error (MPKPE)} (\textit{mm}) as the average error of keypoint position among the sequence.

\begin{table*}[!t]
  \centering
  \caption{
  \textbf{Real-world tracking accuracy on four unseen dancing motions.} For each motion clip, we record both the target and executed joint configurations and compute MPJPE/MPJVE over the entire sequence to evaluate tracking precision and temporal consistency. Remarkably, the real-world performance closely matches the results obtained in simulation, demonstrating that \methodname\ achieves strong zero-shot transfer and maintains high-fidelity whole-body tracking even under real-world dynamics.
  }
  \label{tab:real_dance}
  \resizebox{1.0\linewidth}{!}{
  \setlength{\tabcolsep}{6pt}
  \scriptsize
  \begin{tabular}{lcccccccc}
    \toprule
    \multirow{2}{*}{Backbone} &
    \multicolumn{2}{c}{Can Do Can Go!} &
    \multicolumn{2}{c}{Gokuraku Joudo} &
    \multicolumn{2}{c}{HuoYuanJia/Fearless} &
    \multicolumn{2}{c}{PokerFace} \\
    \cmidrule(lr){2-3} \cmidrule(lr){4-5} \cmidrule(lr){6-7} \cmidrule(lr){8-9}
      & MPJPE ↓ & MPJVE ↓ & MPJPE ↓ & MPJVE ↓ & MPJPE ↓ & MPJVE ↓ & MPJPE ↓ & MPJVE ↓\\
    \midrule
    GMT~\cite{gmt25} & 0.1087 & 1.2560 & 0.1098 & 1.2865 & 0.0921 & 0.7054 & 0.0994 & 0.8217 \\
    TWIST~\cite{twist25} & 0.1253 & 1.1637 & 0.1162 & 1.2731 & 0.1079 & 0.7821 & 0.1047 & 0.8893 \\
    Any2Track~\cite{any2track25} & 0.1039 & 1.1828 & 0.1136 & 1.2366 & 0.0956 & 0.6410 & 0.0928 & 0.8641 \\
    \midrule
    Humanoid-GPT-S  & 0.1024 & 1.0572 & 0.1180 & 1.2362 & \textbf{0.0825} & 0.6209 & 0.0903 & 0.8476 \\
    \rowcolor{linecolor2}Humanoid-GPT-B & \textbf{0.0974} & \textbf{0.9813}  &  \textbf{0.1075} & \textbf{1.2257}  & 0.0858  & \textbf{0.6158} & \textbf{0.0856} & \textbf{0.7325} \\
    \bottomrule
  \end{tabular}
  }
\end{table*}

\paragraph{Results.}
As shown in Table~\ref{tab:sim_scaling_backbone}, the Transformer-based Humanoid-GPT exhibits clear \textit{scaling laws}: enlarging both the motion corpus and the model capacity yields consistent and substantial gains in tracking accuracy and stability. The largest Humanoid-GPT-L model trained on 2B tokens achieves the best performance across nearly all metrics.
MLP and TCN baselines also benefit from scaling but reveal two critical limitations. First, \textit{data scaling saturates}: while larger models eventually reach competitive success rates (e.g., TCN-L achieves 89.05\% at 2B tokens), the gains from 200M to 2B are marginal compared to Humanoid-GPT's continued improvement. Second, \textit{larger models overfit on small data}: when trained on only 2M tokens, MLP-L (75.25\% SR) performs worse than MLP-S (76.89\% SR), and TCN-L (79.85\% SR) underperforms TCN-S (81.48\% SR). This overfitting diminishes with more data, but the MPKPE gap remains significant—even the best baseline (TCN-L at 56.15mm) lags behind Humanoid-GPT-S (43.25mm) by 30\%. These results demonstrate that while MLP/TCN can achieve reasonable success rates with sufficient data, Transformers offer superior tracking precision and more efficient scaling.

\begin{figure}[t]
  \centering
  \includegraphics[width=0.98\linewidth]{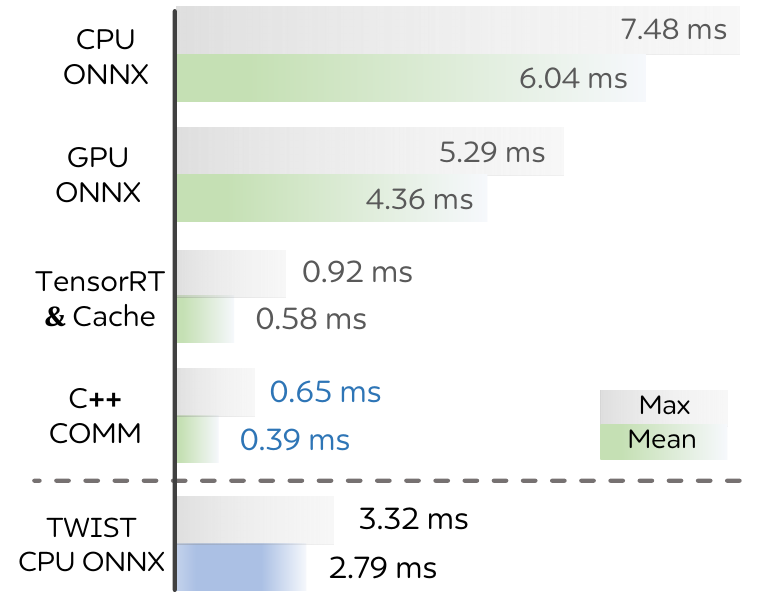}
  \caption{
  \textbf{Comparison of inference latency} among different optimization methods. Our final optimization reaches about 5 times faster than TWIST.
  }
  \label{fig:infer}
  \vspace{-5pt}
\end{figure}

\subsection{Real-world Evaluation}

To further validate the zero-shot generalization of \methodname, we deploy the distilled generalist tracker on the real Unitree-G1 humanoid. Using several pre-recorded dancing sequences, which are entirely unseen during training and consist of highly dynamic full-body motions involving rapid limb coordination and frequent contact transitions. Despite their difficulty, our tracker reproduces these motions in real time without any task-specific fine-tuning, demonstrating strong zero-shot transfer from simulation to the real world. Quantitative motor-sensor analysis in Table~\ref{tab:real_dance} further confirms consistent tracking stability and smooth torque regulation, showing that the physically grounded control learned in simulation effectively transfers to hardware.

We also evaluate \methodname\ in an online whole-body teleoperation setting, where a live MoCap stream is continuously retargeted to the G1’s joint space. Without any additional calibration or adaptation, the tracker directly drives the physical robot to follow the actor’s movements in real time. As shown in Fig.~\ref{fig:real_exp}, the humanoid successfully imitates diverse actions like squatting, stepping, turning, leaning, and expressive arm motions, while maintaining balance and fluid transitions. These results demonstrate unprecedented zero-shot whole-body tracking in the real world, highlighting that motion knowledge distilled from simulation-trained experts and large-scale data seamlessly transfers to embodied execution.

\subsection{Additional Visualization}
\begin{figure*}[t!]
  \centering
  \vspace{-5pt}
  \includegraphics[width=1.0\linewidth]{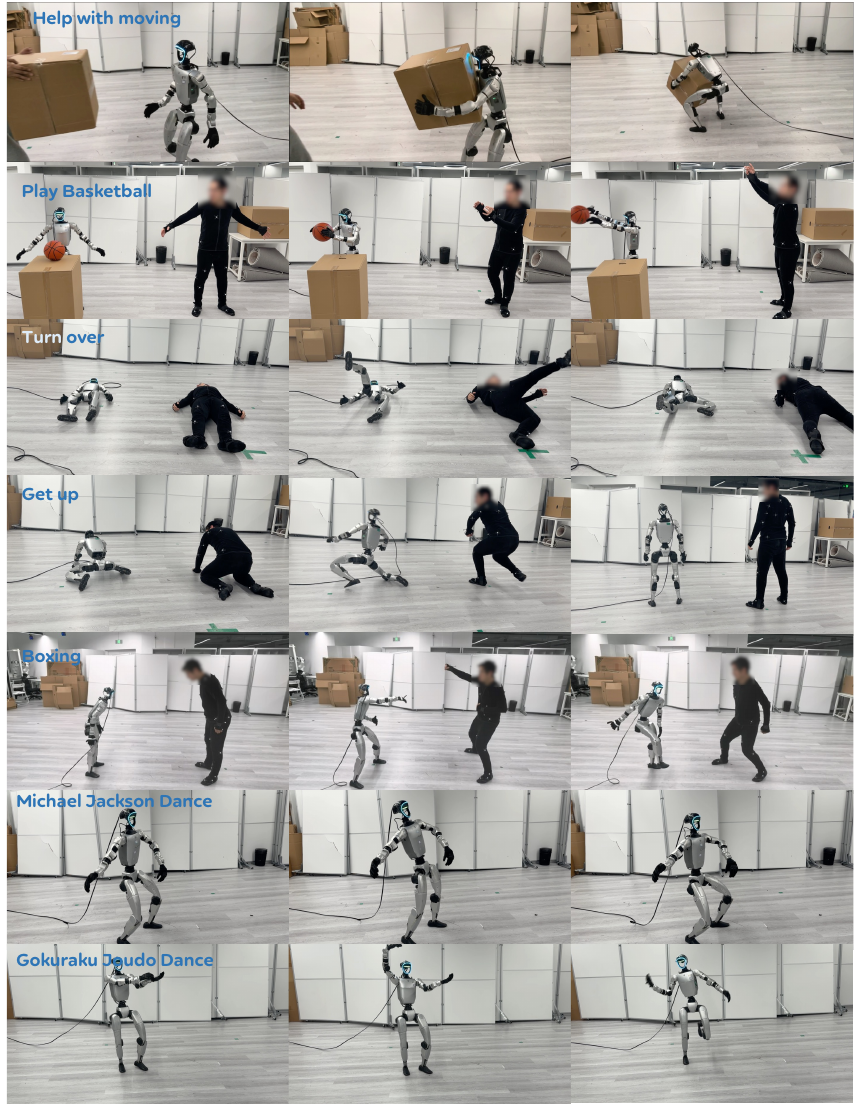}
  \caption{\textbf{Additional Real-world experiments for our Humanoid-GPT.} All motions illustrated are excluded from training to verify generalization capability. Our method can track diverse, complex and high-dynamic motion in a zero-shot manner, especially various dance motions.}
  \vspace{-10pt}
  \label{fig:show1}
\end{figure*}

In this section, we provide additional real-robot examples, as shown in Fig.~\ref{fig:show1}, including both teleoperation demonstrations and zero-shot dancing. As a powerful zero-shot tracker, Humanoid-GPT can execute a wide range of complex behaviors, such as playing basketball, collaboratively carrying boxes with a human partner, and even rolling over and standing up from the ground. We also showcase more iconic dance routines, where motions are directly captured from videos and retargeted to the G1 space; these sequences are not included in our training set.

\subsection{Engineering Optimization}
To satisfy the strict real-time requirements of whole-body humanoid control, we carefully optimized the deployment pipeline to ensure that scaling up the model size does not compromise inference latency. During deployment, the entire model is exported to the ONNX\citep{onnxruntime24} and compiled a compute graph using TensorRT\citep{tensorrt24}. Illustrated in Fig.~\ref{fig:infer}, we also developed a C++-based streaming pipeline to further reduce the communication latency for online teleoperation. These optimizations significantly reduce both computing and memory-access costs. As a result, the final deployed controller achieves an end-to-end inference latency of under \textbf{1.5ms} on a single NVIDIA RTX 4090 GPU. This demonstrates that, despite scaling the model to a substantially larger parameter count, careful engineering and hardware-aware optimization allow the system to maintain real-time performance demanded by full-body humanoid control.

\section{Scaling Laws}
\label{sec:scaling_diversity}

The main paper qualitatively demonstrates monotonic scaling trends.
Here we provide a more formal analysis of scaling laws and quantify the relationship between dataset diversity and generalization.

\begin{figure}[t]
    \centering
    \includegraphics[width=1.0\linewidth]{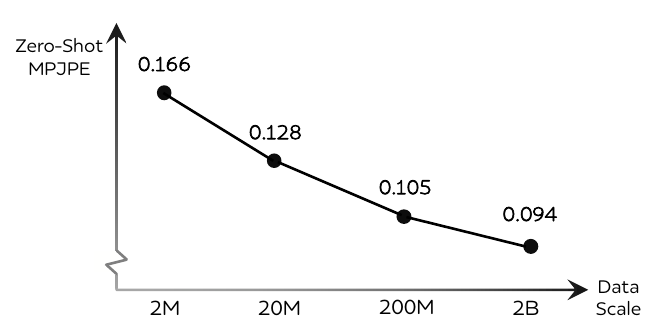}
    \caption{\textbf{Data Scaling up Curve on Zero-shot Performance.}}
    \label{fig:data_scaling}
\end{figure}

\begin{figure}[t]
    \centering
    \includegraphics[width=1.0\linewidth]{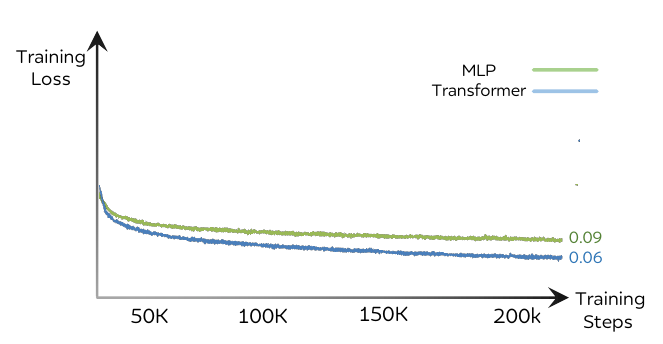}
    \caption{\textbf{Model Scalability Comparison.}}
    \label{fig:model_training}
\end{figure}

\subsection{Data Scaling Up}
\label{subsec:data_scaling}

We vary the number of training tokens $T \in \{2\text{M}, 20\text{M}, 200\text{M}, 2\text{B}\}$ using the Humanoid-GPT-B architecture.
For each $T$, we sample a subset of the 2B-frame corpus without overlap across subsets.
As shown in Fig.~\ref{fig:data_scaling}.
We also observe that the marginal gains decrease slightly between 200M and 2B tokens, suggesting the onset of a data-limited regime for the current model capacity.

\subsection{Model Scaling Ability}
\label{subsec:model_scaling}

We evaluate the scalability of our model by comparing a Transformer-B architecture with an MLP of comparable parameter size, both trained on 2B tokens, with results summarized in Fig.~\ref{fig:model_training}. We observe that the Transformer continues to improve steadily as training progresses, whereas the MLP saturates early, which demonstrates the scalability of Humanoid-GPT.

\begin{figure}[t]
    \centering
    \includegraphics[width=0.9\linewidth]{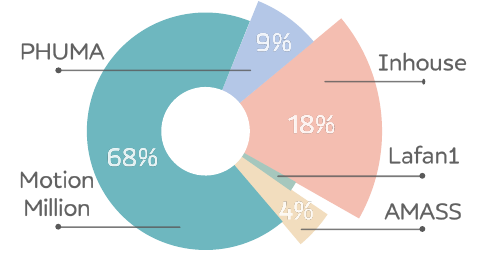}
    \vspace{1pt}
    \caption{\textbf{Data distribution visualization.}}
    \label{fig:data}
\end{figure}

\section{Conclusion \& Future Work}
\label{sec:conclusion}

We introduced \methodname, a GPT-style humanoid motion tracker built by scaling both data and model capacity. Through billion-frame motion curation, clustered expert training, and Transformer-based distillation, our system achieves unified agility, stability, and zero-shot generalization. Experiments in simulation and on the real Unitree-G1 show strong transfer without fine-tuning, enabling reliable real-time whole-body imitation.
Future work includes incorporating richer modalities such as contacts, vision, or language, and extending the framework to interactive or multi-agent scenarios. We also see potential in coupling \methodname\ with longer-horizon planning or VLA-style instruction toward more general-purpose embodied foundation models.

\bibliographystyle{assets/plainnat}
\bibliography{main}

\appendix
\newpage
\clearpage

\section{Summary of Contributions}

\noindent\textbf{Science of Scale:}
We are the first motion tracker with zero-shot ability trained on \textcolor{green}{\textbf{\emph{2B Frame}}} data.
Our training set is over \textcolor{green}{\textbf{$200\times$}} larger than prior trackers.
Scaling by two orders of magnitude requires redesigning the reward and re-tuning key hyperparameters, and it enables the first systematic evidence that video-esti data materially benefits tracking.

\vspace{1.5pt}
\noindent\textbf{Modern Structure:}
We use a scalable causal Transformer for motion tracking. Online tracking cannot access future observations; causal modeling fits this constraint and scales better than MLP and non-causal variants.

\vspace{1.5pt}
\noindent\textbf{Balanced Diversity Matters:}
We introduce HME \textit{Representation Learning} to apply diversity-aware, distribution-balanced sampling in motion tracking. We find that \textcolor{green}{\textbf{diversity and balance}} are both critical for a general tracker.

\section{Additional Ablation Studies}
\label{sec:ablations_pipeline}

\subsection{Number of Experts and Cluster Granularity}
\label{subsec:num_experts}

We next study the effect of the number of motion experts and the granularity of clusters produced by the Harmonic Motion Embedding (HME) representation.
We vary the number of clusters $C \in \{128, 256, 384, 512, 1024\}$ while keeping the total training corpus fixed, leading to different numbers of experts.
For each configuration, we train a distilled Humanoid-GPT-B model on the corresponding expert set and evaluate on the AMASS test split.

As indicated in Fig.~\ref{fig:ablation}, extremely coarse clustering (e.g., 128 experts) leads to experts that cover overly heterogeneous motion patterns, which harms teacher tracking fidelity.
Overly fine granularity (e.g., 1024 experts) increases training cost with conflicting guidelines for the students.
The configuration with roughly $C \approx 384$ experts offers the best balance between diversity, per-cluster coherence, and compute.

\subsection{History Length of Transformer}
Compared with MLPs, Transformers not only offer greater scalability, but more importantly provide substantially enhanced temporal modeling. MLP-based policies typically condition on only a single historical frame, whereas Transformers are inherently designed for sequence modeling. In Fig.~\ref{fig:ablation}, we present the effect of varying sequence length on model performance. All experiments use a Base-sized model with default hyperparameters under a controlled single-factor setting. As shown in the figure, performance continues to improve even with a history of 64 frames. However, due to the quadratic increase in computation with sequence length, we adopt 32 historical frames as the default setting.

\subsection{Environment Number for DAgger Rollout}
\begin{figure*}[t!]
  \centering
  \includegraphics[width=1.0\linewidth]{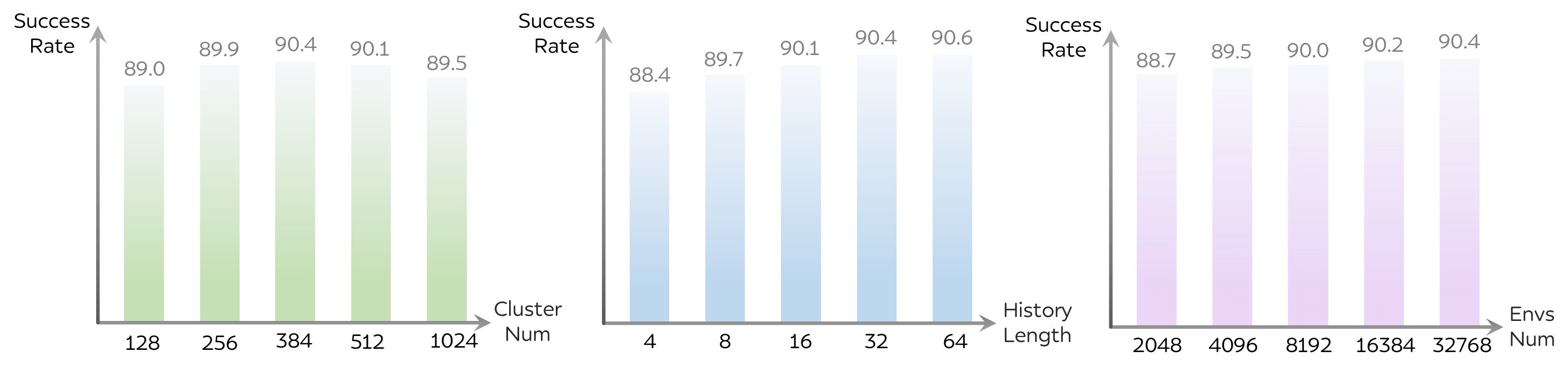}
  \caption{\textbf{Ablation studies for Humanoid-GPT.}}
  \label{fig:ablation}
\end{figure*}

As we scaled up the training data, we found that the number of environments in DAgger also needed to increase accordingly, as shown in Fig.~\ref{fig:ablation}. We ultimately adopted 32K environments. We hypothesize that this is due to the large number of reference motions, where using too few environments may lead to overfitting and forgetting.

\section{Implementation and Reproducibility Details}
\label{sec:implementation}

This section expands the method details that could not fit in the main paper.
We include RL hyperparameters, DAgger schedules, and full compute accounting.

\subsection{RL Expert Training Details}
\label{subsec:rl_details}

\paragraph{Environment and episode setup.}
For each skill cluster, we train a PPO expert in MuJoCo using randomized episode lengths ranging from $T_{\min}$ to $T_{\max}$ frames (typically 600--1200). The control loop runs at 50~Hz, and the reference motion is downsampled to match this frequency. Episodes terminate upon detecting a fall, encountering excessive joint-limit violations, or reaching the time-out horizon. All hyperparameters and reward weights involved in PPO training are listed in Table~\ref{tab:ppo_hyperparams} and Table~\ref{tab:reward_weights}.

To enhance robustness and improve generalization across diverse motion clusters, we apply a set of domain randomization strategies during training. These perturbations are injected into the MuJoCo environment at the beginning of each episode and occasionally throughout rollout generation, ensuring that the learned policy remains stable under variations in dynamics, sensing, and execution conditions. As shown in Table~\ref{tab:dr_items}, we randomize:

\begin{itemize}[leftmargin=1.5em]
\item \textbf{Terrain properties}, including floor friction, maximum terrain height, and procedural noise parameters for terrain generation (noise scale, octaves, persistence, and lacunarity).

\item \textbf{External forces}, where both the interval between force injections and the magnitude of the applied velocity perturbations are sampled from uniform distributions.

\item \textbf{Physical property variations}, including DoF friction scaling, armature scaling, torso center-of-mass shifts, torso mass perturbations, and per-DoF position jittering at the beginning of each rollout.

\end{itemize}

These domain randomization settings are applied uniformly across all PPO expert training environments. For the DAgger stage, we use the same environment configuration and identical randomization scheme, ensuring that the aggregated demonstrations capture the full distribution of dynamic variations encountered by the expert policy.

\begin{table}[t!]
    \centering
    \caption{\textbf{Domain Randomizations.}}
    \label{tab:dr_items}
    \begin{tabular}{cc}
    \toprule
    Item & Random range \\
    \midrule
    \multicolumn{2}{c}{\textbf{Terrains}} \\
    Floor friction & $\mathcal{U}(0.3,\, 2.0)$ \\
    Max terrain height & $0.3$ \\
    Noise scale & $\mathcal{U}(10.0,\, 16.0)$ \\
    Noise octaves & $\mathcal{U}(5.0,\, 8.0)$ \\
    Noise persistence & $\mathcal{U}(0.3,\, 0.5)$ \\
    Noise lacunarity & $\mathcal{U}(2.0,\, 4.0)$ \\
    \midrule
    \multicolumn{2}{c}{\textbf{External Forces}} \\
    Interval range & $\mathcal{U}(5.0,\, 10.0)$ \\
    Velocity magnitude range & $\mathcal{U}(0.1,\, 1.0)$ \\
    \midrule
    \multicolumn{2}{c}{\textbf{Physical Property Changes}} \\
    DoF friction scaling & $\mathcal{U}(0.5,\, 2.0)$ \\
    Armature scaling & $\mathcal{U}(1.0,\, 1.05)$ \\
    Torso CoM position change & $\mathcal{U}(-0.15,\, 0.15)$ \\
    Torso mass change & $\mathcal{U}(-3.0,\, 6.0)$ \\
    Default DoF position jittering & $\mathcal{U}(-0.05,\, 0.05)$ \\
    \bottomrule
    \end{tabular}
\end{table}

\begin{table}[t!]
    \centering
    \caption{\textbf{Hyperparameter settings for training motion experts.}}
    \label{tab:ppo_hyperparams}
    \begin{tabular}{cc}
    \toprule
    Hyperparameter & Value \\
    \midrule
    Env Numbers & 32768 \\
    Batch size & 1024 \\
    Discount factor $\gamma$ & 0.97 \\
    GAE parameter $\lambda$ & 0.95 \\
    Clipping parameter $\epsilon$ & 0.2 \\
    Policy network size & [512, 256, 128] \\
    Critic network size & [512, 256, 128] \\
    Learning rate & $3\times10^{-4}$ \\
    Entropy coefficient & 0.01 \\
    Optimizer & Adam \\
    Training iteration per expert & 3B \\
    \bottomrule
    \end{tabular}
\end{table}

\begin{table}[t!]
    \centering
    \caption{\textbf{Reward weights of different terms.}}
    \label{tab:reward_weights}
    \begin{tabular}{cc}
    \toprule
    Term & Weight \\
    \midrule
    lowerbody keypoints $w_k$ & 1.5 \\
    upperbody keypoints $w_k$ & 0.75 \\
    keypoint position $\alpha_{\text{pos}}$ & 1.0 \\
    keypoint orientation $\alpha_{\text{rot}}$ & 2.0 \\
    keypoint linear velocity $\alpha_{\text{vel}}$ & 0.03 \\
    \bottomrule
    \end{tabular}
\end{table}

\begin{table}[t!]
    \centering
    \caption{\textbf{Hyperparameter settings for DAgger BC.}}
    \label{tab:ppo_hyperparams}
    \begin{tabular}{cc}
    \toprule
    Hyperparameter & Value \\
    \midrule
    Env Numbers & 32768 \\
    Batch size & 32768 \\
    Gradient Clipping & 1.0 \\
    Learning rate & $1\times10^{-4}$ \\
    Num Layers & 12 \\
    Channel dims & 256/384/768 \\
    Optimizer & AdamW \\
    Training iteration & 200k \\
    \bottomrule
    \end{tabular}
\end{table}

\subsection{DAgger Distillation Schedule}
\label{subsec:dagger_details}

We follow a standard DAgger loop for Behaviour Cloning (BC):
\begin{enumerate}[leftmargin=1.5em]
\item Initialize both the expert teacher and the student policy within the simulation environment.
\item At iteration $i$, roll out the student policy and query the expert for the corresponding target action using the same state.
\item Train the student to match the expert’s action, and then update the environment using the student’s executed action.
\end{enumerate}

In practice, we fix the maximal history length $H$ in Eq.~(2) to 32 and maintain history buffers for both the teacher’s actions and the student’s observations. To avoid mode collapse when some experts cover only partial behavior distributions, the batch size used for Behaviour Cloning is kept no smaller than the number of experts.

\begin{table*}[t!]
    \centering
    \caption{\textbf{Approximate compute breakdown.}}
    \label{tab:compute}
    \footnotesize
    \begin{tabular}{lccc}
    
    \toprule
    Stage & Hardware & Total GPU hours & Fraction of total (\%) \\
    \midrule
    PPO experts ($\sim 384$ experts) & RTX 4090 & 12,000 & 75\% \\
    Distillation (Humanoid-GPT-S/B/L) & H100 & 3,000 & 25\% \\
    \midrule
    \textbf{Total} & --- & \textbf{15,000} & \textbf{100}\% \\
    \bottomrule
    \end{tabular}
\end{table*}

\subsection{Compute Cost Breakdown}
\label{subsec:compute_breakdown}

The main paper reports a total compute budget of roughly 15,000 GPU hours. Here we detail the breakdown between expert training and Transformer distillation shown in Table~\ref{tab:compute}.

We emphasize that, once trained, only the distilled Humanoid-GPT policy is required at deployment time.
The expert library is used solely as a training-time teacher and can be discarded afterwards.

\begin{figure*}[t!]
\centering
  \captionof{figure}{\textbf{T-SNE distribution Visualization for our dataset.}}
  \label{fig:tsne}
  \vspace{5pt}
  \includegraphics[width=0.8\linewidth]{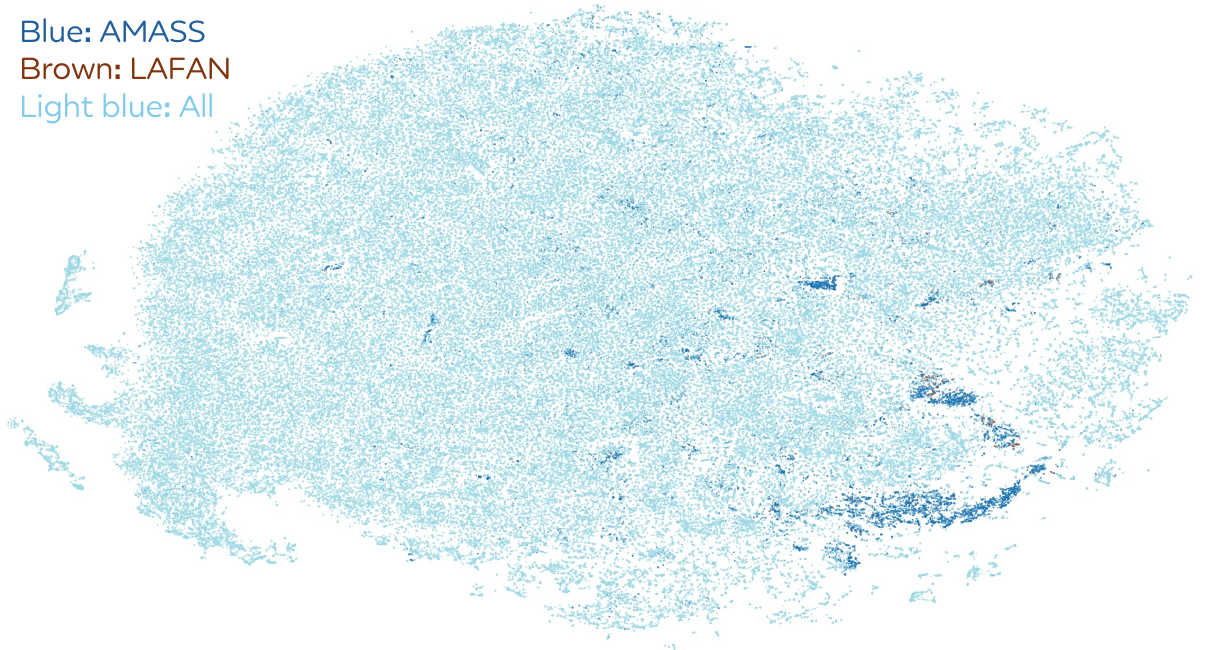}
\end{figure*}

\subsection{T-SNE feature distribution}
We visualize the feature distribution via t-SNE in Fig.~\ref{fig:tsne}. Our data covers a substantially broader region than AMASS+LAFAN1.

\subsection{Deployment Details and Latency Measurements}
\label{subsec:deployment_details}

For completeness, we provide the exact configuration used to obtain the latency results in Fig.~5 of the main paper:
\begin{itemize}[leftmargin=1.5em]
    \item \textbf{Inference hardware:} Single NVIDIA RTX 4090 GPU, CPU: \emph{Intel Core i9-14900KF}.
    \item \textbf{ONNX export:} FP32 weights with CUDA.
    \item \textbf{TensorRT optimization:} Engine built with optimized kernels for causal attention and fused MLPs.
    \item \textbf{Control loop:} End-to-end closed-loop frequency of 50~Hz, including sensor read, inference, PD computation, and actuation commands.
\end{itemize}

The complete deployment stack will be released as configuration files and scripts, enabling reproducible real-time control on G1-like humanoids.

\end{document}